\documentclass[a4paper, 10pt, conference]{ieeeconf}      % Use this line for a4 paper

\IEEEoverridecommandlockouts                              % This command is only needed if 
\usepackage{graphicx}
\usepackage{svg}
\usepackage{amsmath} % assumes amsmath package installed
\usepackage{amssymb}  % assumes amsmath package installed
\usepackage{xcolor}
\usepackage{pifont}
\usepackage{soul}
\soulregister{\cite}{1}
\usepackage{siunitx}
\usepackage{stmaryrd}
\usepackage{algorithm}
\usepackage{algpseudocode}
\usepackage{tabularx}
\usepackage{tikz}
\usetikzlibrary{shapes.geometric}
\usepackage{enumerate}
\let\labelindent\relax
\usepackage{enumitem}
\usepackage{booktabs}
\usepackage{makecell}
\usepackage{multirow}

\usepackage{hyperref} 
\usepackage{cleveref}

\crefname{figure}{Fig.}{Figs.}
\Crefname{figure}{Fig.}{Figs.}

\usepackage[style=ieee, citestyle=numeric-comp, maxcitenames=2, mincitenames=1]{biblatex}
\AtBeginBibliography{\footnotesize}

\newif\ifcenterfigcaptions
\makeatletter
\long\def\@makecaption#1#2{%
\ifx\@captype\@IEEEtablestring%
  \parbox[t]{\hsize}{\footnotesize\noindent #1.~~ #2}%
  \@IEEEtablecaptionsepspace%
\else%
  \@IEEEfigurecaptionsepspace%
  \setbox\@tempboxa\hbox{\footnotesize #1.~~ #2}%
  \ifdim \wd\@tempboxa >\hsize%
    \setbox\@tempboxa\hbox{\footnotesize #1.~~ }%
    \parbox[t]{\hsize}{\footnotesize \noindent\unhbox\@tempboxa#2}%
  \else%
    \ifcenterfigcaptions \hbox to\hsize{\footnotesize\hfil\box\@tempboxa\hfil}%
    \else \hbox to\hsize{\footnotesize\box\@tempboxa\hfil}%
  \fi\fi\fi}
\makeatother

\title{\LARGE \bf
Deployment Is Not Destiny: Robot Recomposition in the Field with Unseen Software, Hardware, and Compute Payloads}

\newcommand{\projectsite}{\href{https://utnuclearrobotics.github.io/deployment_is_not_destiny/}{https://utnuclearrobotics.github.io/deployment\_is\_not\_destiny}}

\author{
Steven Swanbeck\textsuperscript{*},
Jonathan Salfity\textsuperscript{*}, 
Jeffery Gunawan,
Corrie Van Sice,\\ 
Mitch Pryor,
and Robert Blake Anderson
\thanks{\textsuperscript{*}S. Swanbeck and J. Salfity contributed equally to this work.}
\thanks{The authors are with Texas Robotics and the Walker Department of Mechanical Engineering, The University of Texas at Austin, Austin, TX 78712, USA.}
\thanks{Project site: \projectsite. Open-source software will be released at camera-ready submission.}
\thanks{This material is based upon work supported by the University Technology Development Directorate (UTDD), U.S. Army Futures Command (AFC), under Contract No. W911NF-24-C-0067. Any opinions, findings and conclusions or recommendations expressed in this material are those of the author(s) and do not necessarily reflect the views of the Army Futures Command, and no such endorsement should be inferred.
DISTRIBUTION STATEMENT A. Approved for public release; distribution is unlimited. OPSEC \# 10565}
}

\newcommand{\squared}[2][black]{%
  \tikz[baseline=(char.base)]{
    \node[
      shape=rectangle,
      draw=black,
      fill=#1,
      minimum size=1.8ex,
      inner sep=0pt
    ] (char) {\scriptsize #2};
  }
}

\newcommand{\circled}[2][black]{%
  \tikz[baseline=(char.base)]{
    \node[
      shape=circle,
      draw=black,
      fill=#1,
      inner sep=1pt
    ] (char) {\scriptsize #2};
  }
}

\newcommand{\triangled}[3][black]{%
  \tikz[baseline=-0.8ex]{
    \node[
      regular polygon,
      regular polygon sides=3,
      shape border rotate=180,
      draw=black,
      fill=#1,
      minimum size=#2, % <-- older key
      inner sep=0pt
    ] (char) {\scriptsize #3};
  }
}

\newcommand{\diamonded}[3][black]{%
  \tikz[baseline=(char.base)]{
    \node[
      diamond,
      draw=black,
      fill=#1,
      inner sep=1pt,
      xscale=#2
    ] (char) {\scriptsize #3};
  }
}

\definecolor{componentred}{HTML}{FF6666}
\definecolor{componentorange}{HTML}{FFB366}
\definecolor{componentyellow}{HTML}{FFFF66}
\definecolor{componentgreen}{HTML}{66FF66}
\definecolor{componentcyan}{HTML}{66FFFF}
\definecolor{componentblue}{HTML}{99CCFF}
\definecolor{componentpurple}{HTML}{CC99FF}
\definecolor{componentpink}{HTML}{FF99FF}
\definecolor{componentlightpink}{HTML}{FF99CC}
\definecolor{componentteal}{HTML}{66FFB3}
\definecolor{componentrobot1}{HTML}{CCFFE6}
\definecolor{componentrobot2}{HTML}{CCFFFF}
\definecolor{componentrobot3}{HTML}{CCCCFF}

\newcommand{\rba}{\squared[componentrobot1]{1}\!\!}
\newcommand{\rbb}{\squared[componentrobot2]{2}\!\!}
\newcommand{\rbc}{\squared[componentrobot3]{3}\!\!}

\newcommand{\swa}{\circled[componentred]{1}\!\!}
\newcommand{\swb}{\circled[componentorange]{2}\!\!}
\newcommand{\swc}{\circled[componentyellow]{3}\!\!}
\newcommand{\swd}{\circled[componentgreen]{4}\!\!}
\newcommand{\swe}{\circled[componentcyan]{5}\!\!}
\newcommand{\swf}{\circled[componentblue]{6}\!\!}
\newcommand{\swg}{\circled[componentpurple]{7}\!\!}

\newcommand{\hwa}{\triangled[componentred]{1em}{1}\!\!}
\newcommand{\hwb}{\triangled[componentorange]{1em}{2}\!\!}
\newcommand{\hwc}{\triangled[componentyellow]{1em}{3}\!\!}
\newcommand{\hwd}{\triangled[componentgreen]{1em}{4}\!\!}
\newcommand{\hwe}{\triangled[componentpink]{1em}{5}\!\!}
\newcommand{\hwf}{\triangled[componentlightpink]{1em}{6}\!\!}

\newcommand{\cpa}{\diamonded[componentteal]{0.7}{1}\!\!}

\newcommand{\Unseen}{Nondum Visum }
\newcommand{\unseen}{nondum visum }

\definecolor{hlred}{RGB}{255, 200, 200}

\definecolor{hlgreen}{RGB}{180, 255, 160} 
\begin{document}

\maketitle
\thispagestyle{empty}
\pagestyle{empty}

%%%%%%%%%%%%%%%%%%%%%%%%%%%%%%%%%%%%%%%%%%%%%%%%%%%%%%%%%%%%%%%%%%%%%%%%%%%%%%%%
\begin{abstract}

The tight coupling of subsystems in most robots, though a natural consequence of their complexity, leads to monolithic designs that are time-consuming and difficult to adapt after initial deployment. 
To address this challenge, we present a framework and supporting abstractions for recomposition during runtime that enable robots to quickly integrate previously unseen modular software, hardware, and compute payloads. 
Our approach allows non-expert users to quickly add new capabilities in the field through a true plug-and-play process. 
Crucially, new resources are not only immediately available to a host robot but are also shared with distributed peers, enabling compute-constrained systems to access powerful new remote capabilities. 
Our framework reduces reconfiguration time to a matter of minutes with no developer intervention, in stark contrast to the hours of expert effort often required for traditional manual integration. 
We demonstrate our method in two disaster response scenarios, including radioactive source localization at an operational nuclear reactor facility and a thermal-guided search for people in dark, difficult-to-reach spaces. 
These demonstrations show how in-field recomposition provides timely, flexible, and accessible adaptation to dynamic requirements, representing a critical step toward creating robots that can quickly evolve alongside the tasks, technologies, and environments they support.
% We release our framework open source\footnote{Link will be provided in the final, unanonymized manuscript.}.

\end{abstract}
\section{Introduction}\label{sec:introduction}

Most robots are designed and deployed as static systems preconfigured for a narrow set of tasks and environments.
This is a consequence of complexity: robots are composed of densely integrated software, electrical, and mechanical subsystems that must be engineered to operate in concert. 
The result is purpose-built robots that are hardened through in situ testing, often deployed with fixed configurations for their entire operational lives \cite{citizen_developer_framework_2022, component_based_robot_engineering_p1_2009, robot_software_reconfiguration_2025}. 
Aside from occasional software updates, there is limited opportunity to extend or modify their functionality once they are in the field.

This paradigm produces reliable and high-performing systems, but it also imposes rigidity \cite{adaptive_robotics_2022}. 
Robots may quickly become ineffective when mission requirements change or new technologies become available, and reconfiguration for a new purpose often requires the intervention of experts. 
In response, some researchers and manufacturers pursue general-purpose robots initialized with a wide range of capabilities to perform diverse tasks without modification \cite{general_purpose_fms_2023}. 
However, such systems are frequently overbuilt relative to the tasks they perform, increasing cost and complexity while still failing to adapt to unanticipated future mission demands. 

\begin{figure}[!t]
    \centering
    \vspace{1ex}
    \includegraphics[width=\columnwidth]{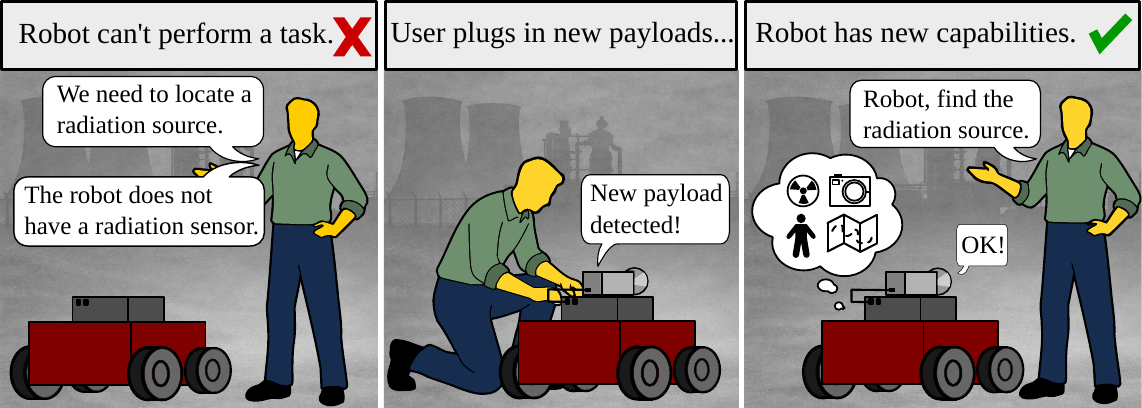}
    \caption{\textbf{Core Concept:}
    Our framework allows robots to be quickly modified with plug-and-play \textit{\unseen payloads} for ad hoc tasks.
    Non-roboticist end-users can quickly reconfigure a robot in the field without reprogramming, and robots require no prior knowledge of introduced payloads.
    % Robots initially not equipped with domain specialized payloads can be quickly modified by a non-roboticist end-user within minutes.
    % An end user is able to quickly modify a robot with plug-and-play \textit{\unseen payloads} of which it has no prior knowledge.  
    Within minutes, a robot can compose new capabilities into goal-directed task plans.} 
    % The end user needs no expertise in robot programming to dynamically modify the robot in the field.
    % }
    \label{fig:concept}
    \vspace{-15pt}
\end{figure}

Modularity offers an alternative to this paradigm \cite{trends_in_reconfigurable_modular_robots_2017}. 
Rather than attempting to anticipate every possible requirement, modular robots can be reconfigured by adding and removing hardware and software modules as needed. 
Research has explored modularity for extending  functionality and deploying new payloads in the field, but these systems often fall short of enabling open-ended runtime adaptation. 
The reason for this is twofold: (i) they typically require a priori knowledge of modules, leading to a fixed modular configuration space \cite{plug_and_produce_collaborative_Schou_2017, concert_rossini_2026}, and (ii) integrating new capabilities often requires a reboot or developer intervention, slowing deployment and limiting responsiveness to unforeseen tasks \cite{reconfigurable_software_2002, heuss_extendable_2022}.

% \newcommand{\ymark}{\ding{51}}
% \newcommand{\pmark}{\ding{109}}
% % \newcommand{\nmark}{\ding{55}}
% \newcommand{\nmark}{}

\newcommand{\ymark}{$\bullet$}
\newcommand{\pmark}{$\circ$}
\newcommand{\nmark}{$\times$}

\begin{table*}[!t]
\centering
\caption{Positioning of this work against the related-work threads of \Cref{sec:related_work}.
\ymark~= full support, \pmark~= partial support, \nmark~= not supported; further details are provided in \Cref{sec:related_work}.
No prior thread achieves full support across every column; ours is the only row of all \ymark.}
\label{tab:positioning}
\footnotesize
\setlength{\tabcolsep}{2pt}

\newcommand{\full}{\ymark}
\newcommand{\partialc}[1]{\pmark}
\newcommand{\nonec}{\nmark}
\newcommand{\nonecnote}[1]{\nmark\rlap{$^{#1}$}}

\begin{tabular*}{\textwidth}{@{\extracolsep{\fill}}l l c c c c c c c c@{}}
\toprule
    % \makecell[l]{\textbf{Research Thread}}
    & \multicolumn{3}{c}{\textbf{Limitations of Prior Work}}
    & \multicolumn{3}{c}{\textbf{Additional Key Properties}}
    & \multicolumn{3}{c}{\textbf{Component Types}} \\
\cmidrule(lr){2-4} \cmidrule(lr){5-7} \cmidrule(lr){8-10}
\makecell[l]{\textbf{Research Thread}\\Representative or summary work}
    & Configuration space 
    & \makecell{No\\reboot} 
    & \makecell{No manual\\intervention}
    & \makecell{Self-contained\\payloads} 
    & \makecell{Auto.\\shared} & \makecell{Auto.\\composed}
    & Hardware 
    & Software 
    & Compute\ \\
\midrule

% Software/ middleware
Composition middleware \scriptsize\cite{ros2_2022,citizen_developer_framework_2022,xbot2_middleware_2023,murt_middleware_2023,coral_2026}
    & Indicated processes 
    & \partialc{} 
    & \nonec
    & \nonec 
    & \partialc{} 
    & \nonec
    & \nonec 
    & \full 
    & \partialc{1} \\
Software reconfiguration \scriptsize\cite{reconfigurable_software_2002,dynamic_software_updates_2023,software_variability_2023,robot_software_reconfiguration_2025}
    & Parameters 
    & \partialc{} 
    & \partialc{}
    & \nonec 
    & \nonec 
    & \nonec
    & \nonec 
    & \full 
    & \nonec \\
Adaptive resource management \scriptsize\cite{temoto_2022}
    & Known resources 
    & \full 
    & \full
    & \nonec 
    & \full 
    & \nonec
    & \partialc{} 
    & \full 
    & \nonec \\
    % & \partialc{1} \\
Distributed \& cloud/fog robotics \scriptsize\cite{pdra_2020,cloud_robotics_resources_2021,fogros2_2023}
    & Known services 
    & \partialc{}
    & \partialc{}
    & \nonec 
    & \partialc{} 
    & \nonec
    & \nonec 
    & \partialc{} 
    & \full \\

% Robots
Modular robots \scriptsize\cite{concert_rossini_2026,Chennareddy_Agrawal_Karuppiah_2017,Liang_Wu_Tu_Lam_2024,thorvald_Grimstad_From_2017,mars_Xu_Li_2022,Guri2024HeftyAM,modkom_Wiedemann_2024,Modular_reconfigurable_mobile_robotics_2012}
    & Closed module set 
    & \partialc{} 
    & \partialc{}
    & \partialc{} 
    & \nonec 
    & \partialc{}
    & \full 
    & \partialc{} 
    & \nonec \\
    
Manufacturing \scriptsize\cite{Arai2000PlugProduce,plug_and_produce_collaborative_Schou_2017,Generic_plug_and_produce_Profanter_2021,heuss_modular_2019,heuss_extendable_2022}
    & Device database 
    & \full 
    & \partialc{}
    & \partialc{} 
    & \nonec 
    & \nonec
    & \full 
    & \partialc{} 
    & \nonec \\

\midrule
\textbf{Ours}
    & \textbf{Open module set} 
    & \full 
    & \full
    & \full 
    & \full 
    & \full
    & \full 
    & \full 
    & \full \\
\bottomrule
\end{tabular*}

\vspace*{-5pt}
% \vspace{3pt}
% \footnotesize
% \textbf{Structural gaps:}\quad
% $^{1}$Requires a priori known, registered payloads\quad
% $^{2}$requires a priori known/registered payloads\quad
% $^{2}$Self-contained but payload must supply its own power/compute rather than drawing on host resources\quad
% $^{3b}$requires driver/support software already installed on the host before the payload can be used\quad
% $^{3}$Resources are scheduled/selected, not composed into goal-directed task plans
\end{table*}

The main contribution of this paper is an implemented framework for open-ended runtime modularity supporting the integration of \emph{\unseen payloads}: self-contained software, hardware, or compute resources that were not known by the robot--or that may not have even existed--at the time of original deployment. 
Our contribution is divided into three parts. 
First, we introduce composable abstractions that allow \unseen payloads to be integrated as modular \emph{components}--the atomic building blocks of our system--regardless of the functionality they provide.
% \hlgreen{While we describe abstractions for software, hardware, and compute resources, we place particular emphasis on software composition, as hardware interfaces are typically more constrained by physical and electrical standards, whereas software can evolve far more rapidly due to its substantially lower modification cost \cite{robotics_software_review_2024}.}
Second, we introduce a complimentary framework that autonomously discovers and manages payloads introduced or removed during runtime and composes capabilities afforded by components into goal-directed task plans, supporting recomposition across multiple layers of the robotics stack without developer intervention.
Finally, we validate this approach with two demonstrations in application domains that require timely adaptation using three host robots and a diverse set of payloads for perception, actuation, human-robot interaction, and external compute. 
Our scope focuses specifically on the scheduling, information flow, and utilization of newly introduced payloads.
Payload selection for a given environment and task
% , and payload physical placement (in the case of hardware) 
remains the responsibility of a human teammate.

% \blue{[trying to nail down on behaviorial software scope statement]}
% We are principally concerned with the scheduling, flow of information between processes, and utilization of new payloads; verification of correctness with regards to factors including hardware placement or fitness with respect to environmental or task factors remain the responsibility of a user. That is, selection of the components and factors outside of their software configuration (including physical placement in the robot) must be manually performed. 
% Our approach shifts robots from fixed-function tools into versatile, reusable systems, not by anticipating every possible use case at design time, but by enabling rapid recomposition during runtime. 
% Through the plug-and-play integration of \unseen payloads, robots can be transformed into ad hoc specialists for the current task before quickly being reconfigured for the next.
% \Cref{fig:concept} captures this core concept graphically.
% This capability extends operational lifespan and allows robots to adapt to emerging and unforeseen mission requirements in dynamic environments.

\section{Related Work}\label{sec:related_work}

As summarized in \Cref{tab:positioning}, prior work clusters into two broad groups: middleware, software, and runtime resource-management frameworks (\Cref{sec:related_work_software}), and modular and reconfigurable robots (\Cref{sec:related_work_hardware}). 
Across both groups, the same two limitations reoccur: (i) a closed, a priori known configuration space and (ii) integration that requires a reboot or developer intervention.
% These limitations motivate our development of a holistic framework that enables plug-and-play recomposition during runtime with previously unseen software, hardware, and compute payloads.

\subsection{Middleware, Software, and Resource Management}\label{sec:related_work_software}
Middleware provides one of the most pervasive abstractions for integrating diverse
software components, offering interfaces between otherwise decoupled processes.
ROS \cite{ros2_2022} remains the most widely used middleware, but many alternatives improve accessibility to non-experts \cite{citizen_developer_framework_2022}, provide hardware-layer abstraction \cite{xbot2_middleware_2023}, or support execution across compute-constrained devices \cite{murt_middleware_2023}.
These middlewares standardize communication between components, but do not consider higher-level scheduling or interactions between components and introducing a new component requires a developer action and, in most cases, relaunching the affected processes.
We extend the abstraction layer in \cite{coral_2026}, our own prior work, which enables complex software systems to be composed from independent components that have no knowledge of each other prior to runtime but does not automatically respond when resources are added or removed during runtime.

While software reconfiguration has been studied for decades \cite{reconfigurable_software_2002,software_variability_2023}, there continues to be a significant gap between the state of the art and the state of practice \cite{robot_software_reconfiguration_2025}.
Only parameter-level reconfiguration is widely used in practice, with runtime reconfiguration via integration of independently executed components remaining an open challenge \cite{robot_software_reconfiguration_2025}.
Orthogonal work addresses safely applying system-wide updates to running software but explicitly rejects the kind of partial updates required in the reconfigurable robotics setting \cite{dynamic_software_updates_2023}.
\cite{temoto_2022} shares many of our motivations, introducing a fault-tolerant framework for dynamic allocation of robot components that may be added, removed, or fail during runtime, and is the only prior thread to share physical resources such as sensors automatically across robots, but components must be known in advance and fault-tolerant behaviors are manually programmed rather than being generated automatically.
Distributed, cloud, and fog robotics work \cite{pdra_2020,cloud_robotics_resources_2021,fogros2_2023} instead offloads computation to peers or the cloud.
\cite{pdra_2020} allocates tasks automatically, but only over a software network whose tasks and hosted resources are known to all robots in advance.
\cite{fogros2_2023} provisions and populates networked cloud machines on demand, but the decision of which components to offload is specified by the developer in a launch description.
In both cases, allocation schedules known capabilities rather than composing newly discovered ones into task plans and the software required to use a given resource must already be integrated and known by the robot.
 
\subsection{Modular and Reconfigurable Robots}\label{sec:related_work_hardware}
Prior work in modular robotics has focused heavily on hardware modularity, including standardized mechanical and electrical interfaces that enable physical reconfiguration \cite{Modular_reconfigurable_mobile_robotics_2012,Chennareddy_Agrawal_Karuppiah_2017,Liang_Wu_Tu_Lam_2024}.
Resulting modular robot application domains range across agriculture \cite{thorvald_Grimstad_From_2017,Guri2024HeftyAM,mars_Xu_Li_2022}, space \cite{modkom_Wiedemann_2024}, and construction \cite{concert_rossini_2026}. 
Common limitations are that these systems reconfigure among a fixed, pre-engineered kit of modules implying reconfiguration cannot incorporate an unknown module without developer intervention.

Manufacturing settings emphasize similar runtime flexibility as we are proposing.
Skill-based software frameworks \cite{heuss_modular_2019,heuss_extendable_2022} enable modular composition of robot capabilities, while \cite{plug_and_produce_collaborative_Schou_2017} and \cite{Generic_plug_and_produce_Profanter_2021} automatically detect incoming hardware modules. 
However, \cite{heuss_modular_2019,heuss_extendable_2022,plug_and_produce_collaborative_Schou_2017} rely on pre-defined skill catalogs or static semantic knowledge bases, preventing the seamless integration of previously unknown payloads without developer intervention. 
\cite{Generic_plug_and_produce_Profanter_2021} allows self-describing components, at the cost of requiring each component to supply its own power, compute, and network connection, which prevents utilization of superior host resources and requires payloads to be custom-engineered for compatibility rather than integrating off-the-shelf modules.

\begin{figure*}[htbp!]
    \centering
    \vspace*{1ex}
    \includegraphics[width=0.95\textwidth]{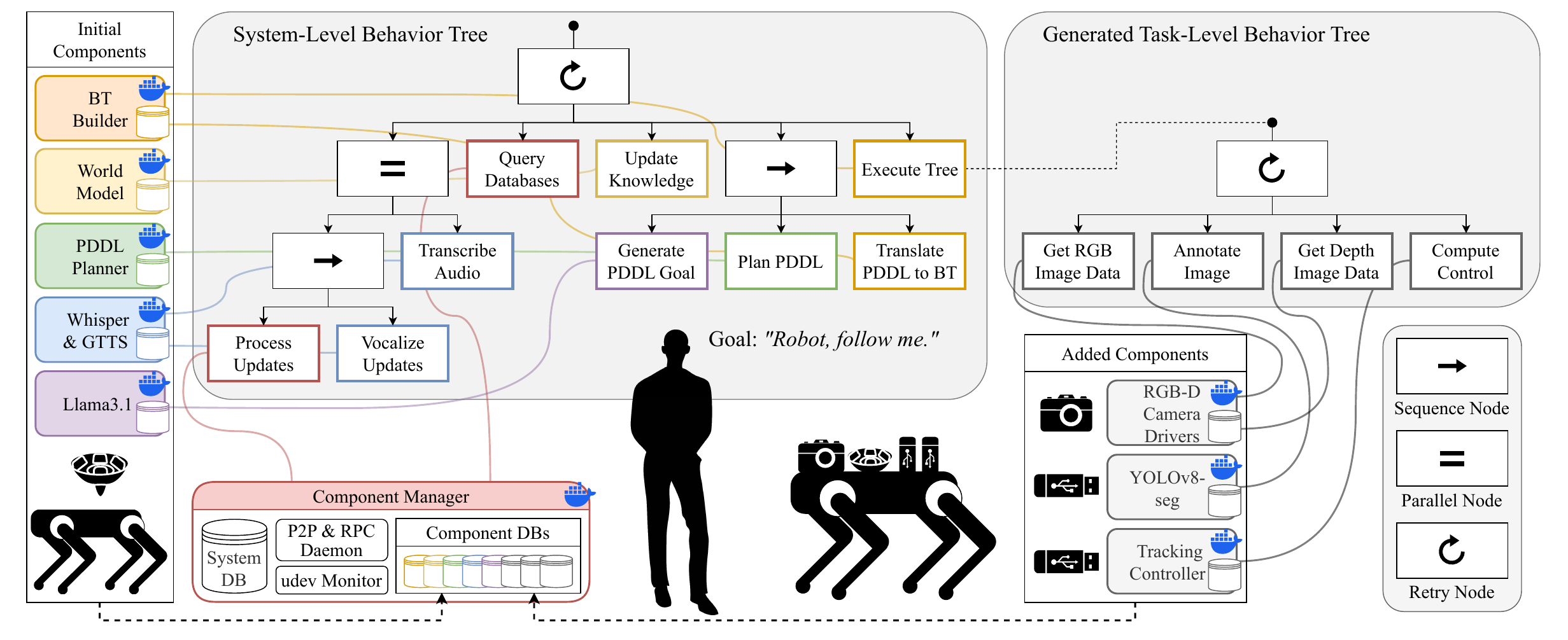}
    % \caption{An overview of our modular architecture. Each software component is individually containerized and features a special relational database storing compiled libraries and configuration and planning files that allow its capabilities to be used by the host robot or distributed peers. A special \texttt{Component Manager} component stores the databases of all running components, monitors firing of udev events to detect newly connected software and hardware components, runs a peer-to-peer server that manages connections with distributed peers, and maintains a system-level database. A fixed system-level behavior tree is engineered by the user that leverages behaviors afforded by the components available before runtime. Behaviors afforded by newly integrated software and hardware components are incorporated into the system's skill repository, from which task-level behavior trees are generated based on provided natural language inputs. Behaviors made available by each component are indicated with color and arrows showing origin.}
    \caption{\textbf{Framework Overview:} The system begins with a set of software, hardware, and compute \textit{initial components}. 
    Each software component affords a set of behaviors that are composed into a \textit{system-level behavior tree} responsible for processing system updates and task-planning. 
    A dedicated \textit{component manager} is responsible for discovery and management of \unseen \textit{added components} introduced during runtime. 
    When provided a task goal, the robot leverages available components--including those introduced during runtime--to generate and execute a \textit{task-level behavior tree}. 
    Behaviors afforded by each component are indicated with color and wires showing origin.}
    \label{fig:system_diagram}
    \vspace{-10pt}
\end{figure*}
\section{Approach}\label{sec:approach}
To enable integration of novel resources and adaptation to changing mission requirements, our framework is built around the principle of \emph{compositionality} \cite{compositional_thinking_2022}, where every \unseen payload is treated as a self-contained component that can be dynamically integrated into the robot's decision-making and execution pipeline. 
Existing low-level interfaces for each component type--software, hardware, and compute--are unified and extended where required to produce similar composable properties (\Cref{subsec:approach:component_design_abstractions}).
A dedicated \texttt{Component Manager} combines these interfaces into a top-level abstraction that handles component discovery and utilization (\Cref{subsec:approach:component_manager}).
The result is a framework that supports continuous recomposition with \unseen payloads during runtime in response to evolving mission requirements with no developer intervention and minimal system downtime.

\subsection{Component Design and Abstractions}\label{subsec:approach:component_design_abstractions}

Recomposition with \unseen payloads requires abstractions that promote composability from low-level discovery and installation to high-level task planning and execution. 
Each component type must be handled differently, requiring unique abstractions that produce similar composable properties.

Hardware abstraction is relatively straightforward: interfaces between devices are already highly standardized through physical connectors and signal protocols, and operating systems already support device enumeration and access through their device management stacks. 
To leverage preexisting interfaces, our hardware abstraction handles devices connected via USB.
Hardware is physically mounted to host robots using a mounting plate with a rail-based mechanism for fast recomposition of larger components and a 3D printed vise specifically for forward-facing camera components.
% These mounts are not currently proprioceptive, so the pose of each component relative to the robot must be manually specified by a human teammate.
% As a result, checks on proper hardware positioning (i.e. positioning a camera such that it can see in front of the robot) are not 

Preexisting distributed compute interfaces are also mature. 
Ethernet, WiFi, and IP provide well-defined mechanisms for communication between distributed peers.
In our abstraction, peers connect via Ethernet or WiFi on a predetermined subnet, perform peer discovery via UDP multicasting, and transmit data to known peers over TCP. 
Each compute component runs the same peer-to-peer (P2P) and remote procedure call (RPC) daemon that performs these functions.

Software is relatively underconstrained and thus requires the most added structure. 
We extend the abstraction in \cite{coral_2026}, which coordinates control between Docker-containerized \cite{docker_2014} processes with behavior trees \cite{bts_collendanchise_2016} using ROS \cite{ros2_2022} for interprocess communication.
While \cite{coral_2026} improves compositionality with containers that can run reliably across systems and expose functionality via behaviors and interfaces, it is unable to support a dynamic set of components during runtime, sharing of software resources with distributed peers, or automatic composition of behaviors into task plans. 
To overcome these limitations, our abstraction requires software components to provide (i) \cite{coral_2026}-compliant compiled images and (ii) a corresponding keyword-indexed relational database containing low-level compiled behavior and interface libraries for various CPU architectures and high-level PDDL \cite{pddl_1998} actions and domain and problem fragments corresponding to the new payload's afforded behaviors.
This extension provides two benefits. 
Firstly, containerized processes that serve behaviors are reliably run across hosts regardless of operating systems or dependencies. 
Secondly, relational databases allow easy sharing of low-level behavior and interface libraries and high-level planning definitions between systems.
% enabling utilization of resources afforded by software payloads without prior knowledge of their capabilities or internal implementations. 
% New software components are introduced via USB flash drives plugged into a host computer and discovered via triggered \texttt{udev} events. \redtext{do we need to say this last line? it's very clearly captured in III B 1}

By unifying or extending existing interfaces and abstractions for software, hardware, and compute, we create a system of modular components that can be reliably discovered and utilized during runtime. 
Management of these components during operation is performed by the \texttt{Component Manager}, which is described in the following subsection.

\subsection{Component Discovery and Management During Runtime}\label{subsec:approach:component_manager}
% \blue{component manager does not manage payload to payload or payload to host spatial relationships, i.e. component to component spatial relationships.
% This is a current responsibility of the user.}
We implement discovery and management of \unseen payloads in a \texttt{Component Manager} which robots and external compute components must run to participate in our system.
The \texttt{Component Manager} is responsible for managing the lifetimes of all other available components and reacts to local system updates of three types:
\begin{enumerate}
    \item \textbf{USB Block Device Updates:} Added or removed USB block devices are detected by fired \texttt{udev} events on the local system.
    When new USB block devices are plugged in, they are scanned for any valid software components, which are marked by a compiled image and relational database, as previously described.
    Before loading new software, the host verifies images are compiled for its CPU architecture and that it has sufficient free disk space to transfer the images and GPU memory to run any GPU-intensive processes. Once these checks have passed, the images for all found components are loaded and the database files are copied to the host.
    % The images for all found components are loaded and the database files are copied to the host.
    A container is instantiated from the image after being configured with system-specific networking settings and launched using \cite{coral_2026}.
    Once active, the behavior and interface libraries stored in the corresponding component database can be used to interact with the processes running inside the component container. Upon removal, all associated containers are stopped, images are removed, and databases and extracted contents are pruned from the system.
    \item \textbf{New Device Nodes:} Non-block USB devices associated with sensors, actuators, or other devices are automatically mapped into containers associated with concurrently introduced software.
    This enables plug-and-play integration of physical hardware that can be reliably accessed by its corresponding driver software.
    \item \textbf{Peer Signaling:} To extend to distributed systems, a peer discovery and RPC daemon watches for new networked peers. When discovered, component databases are exchanged, allowing peers to generate and execute plans that leverage each other’s capabilities. Liveness checks ensure pruning upon peer disconnection. Databases from newly introduced components are automatically shared with known peers and alerts are issued to peers when local components are removed to prevent others from using no-longer-available resources.
\end{enumerate}
% All this information, including relations between devices and data, is stored in a system-level database that is queried to share resources with peers and prune contents associated with removed components from the system.

\begin{figure*}[htbp!]
    \centering
    \vspace*{1ex}
    \includegraphics[width=0.95\textwidth]{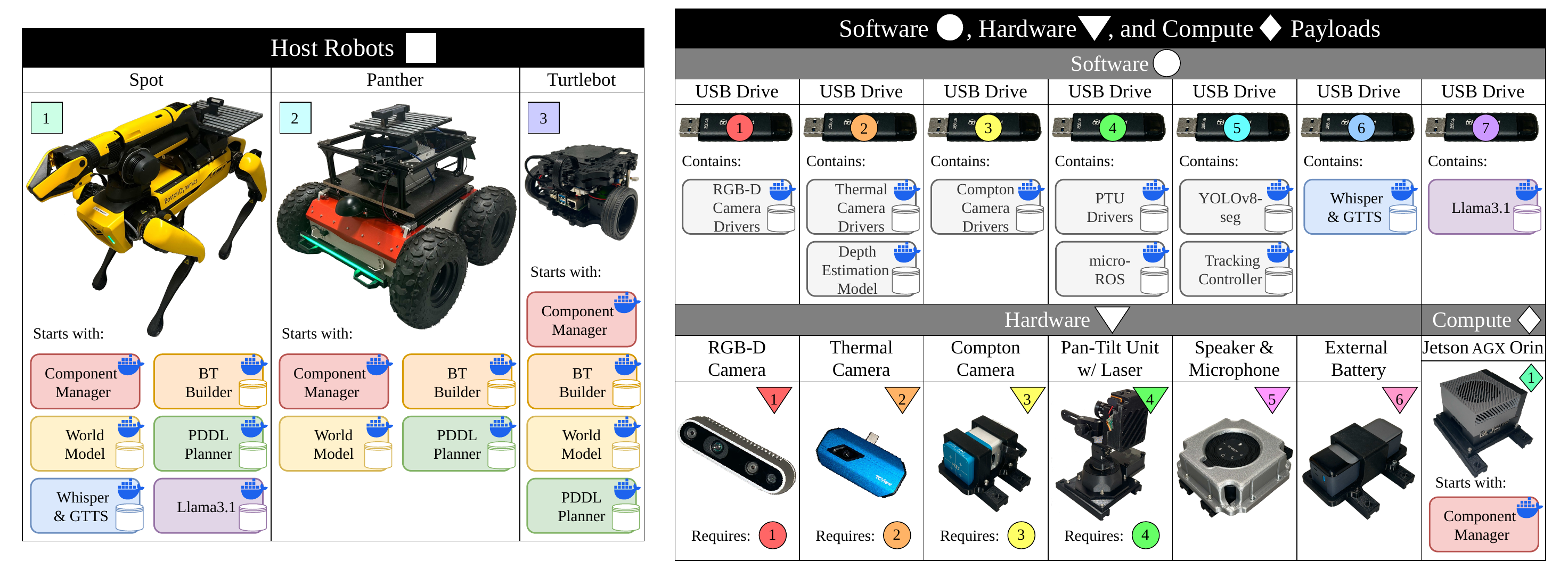}
    % \caption{The set of host robots and software, hardware, and compute payloads used in our demonstrations. 
    % Each host robot and the external compute payload starts with a subset of initial software components from Fig. \ref{fig:system_diagram}. 
    % Software is injected into host robots during operation via USB drives, with each containing one or several software components. 
    % Hardware payloads are interfaced during operation via specialized mounts on each host system. 
    % Software drivers required to use hardware payloads are indicated. All payloads have a unique identifying tag used to reference them in Fig. \ref{fig:netl_demo} and Fig. \ref{fig:ahg_demo}.}
    \caption{\textbf{Host Robots and \Unseen Payloads:} 
    Each host robot and the external compute payload starts with a subset of initial software components from \Cref{fig:system_diagram}. 
    Software is injected into host robots during operation via USB drives, with each containing one or several software components. 
    Hardware payloads are mounted during operation via a rail-based mounting system.
    Software payloads required to use hardware payloads are indicated. 
    Host robots and payloads have unique colored identifying tags used throughout \Cref{sec:demonstrations}, \Cref{fig:netl_demo}, \Cref{fig:ahg_demo}, and \Cref{fig:outcomes}.}
    \label{fig:components}
    \vspace*{-10pt}
\end{figure*}

\subsection{Component Utilization}
These abstractions and their handling by the \texttt{Component Manager} allow \unseen payloads to be introduced and removed reliably during operation. 
The flows of logical control and information between components are coordinated by a \textit{system-level behavior tree} composed of behaviors afforded by the software components available before runtime. 
The structure of this tree can be altered depending on the desired system behavior and available components, but it typically includes mechanisms for component monitoring, updating a world model for planning, handling task-level planning requests, and communicating with a human teammate. 
The system-level behavior tree used in our demonstrations is shown in \Cref{fig:system_diagram}.
When executing this system-level behavior tree, a human teammate provides task goals.
In this work, we use a speaker and microphone hardware component and a software component running a local speech-to-text model \cite{whisper_2023} and text-to-speech API to enable natural language-based interactions and goal specifications. 
We use PDDL to generate task plans and maintain a world model with actions, predicates, and objects amalgamated from all component databases. 
Natural language requests are translated into aligned PDDL goals using a software component running a local large language model \cite{llama3_2024}.
The databases include mappings between PDDL actions and behaviors afforded by running components, allowing a generated task plan to be translated into an equivalent \textit{task-level behavior tree} using an approach similar to \cite{plansys2_2021} that is dynamically inserted and run as part of the system-level behavior tree, as shown in \Cref{fig:system_diagram}.

By monitoring for updates from the local \texttt{Component Manager} inside the system-level tree and re-querying all stored component databases for planning definitions and behavior and interface libraries each time an update is processed, we can continue to generate and execute plans that use the current dynamic set of available components, including those shared from distributed peers.

\subsection{The Result: Plug-and-Play System Recomposition}
The central outcome of our approach is the ability to dynamically recompose a robot using \unseen payloads, whether physically attached or discovered over a network, to rapidly adapt to new tasks.
% Because our framework treats software, hardware, and compute compute as compatible payloads, the associated logic, morphology, and computational power can be dramatically improved at runtime.
% While hardware payloads are necessarily constrained by the host’s volumetric footprint and I/O ports, network-based capability sharing allows the robot's access to compute and software to be bound only by the resources available over its network and the quality of the network itself.
We intend our framework to enable end-users to simply (i) plug in new payloads, which are automatically discovered, installed, and shared to become immediately useful; and (ii) issue high-level goals that are converted into executable task plans composed of capabilities afforded by available components.
\Cref{sec:demonstrations} demonstrates this approach in real-world scenarios.
\section{Demonstrations}\label{sec:demonstrations}

We present demonstrations spanning three different host robots and a diverse set of \unseen payloads to validate our framework's ability to support recomposition during runtime.
Our demonstrations illustrate how the abstractions and framework introduced in \Cref{sec:approach} enable rapid adaptation to changing mission requirements and environments, especially in scenarios where required capabilities cannot be fully anticipated prior to deployment.
They collectively demonstrate four key attributes that support plug-and-play recomposition during runtime:

\begin{enumerate}[label={\textbf{A\arabic*:}}, ref={\textbf{A\arabic*}}]
    \item \label{attribute:A} \textbf{Incremental Payload Integration}: Adding and removing payloads to provide new capabilities as mission needs evolve.
    
    \item \label{attribute:B} \textbf{Distributed Capability Sharing}: 
    Enabling robots to share newly introduced local payloads and utilize remote payloads hosted on another peer system.
    
    \item \label{attribute:C} \textbf{Host-Agnostic Integration}: Reusing the same payloads across different host robots, tasks, and environments.

    \item \label{attribute:D} \textbf{The Concert of Independently Designed Payloads}: 
    Cooperation between independently designed \unseen payloads to achieve a provided task goal.
\end{enumerate}

The host robots and payloads from our demonstrations are shown in \Cref{fig:components}, which also provides identifying tags used throughout this section.
Each of the host robots has a unique computational profile:
\begin{itemize}
\item \textbf{Boston Dynamics Spot} host robot \rba\ in Demonstration 1 with an NVIDIA Jetson AGX Orin with \SI{64}{\giga\byte} shared RAM and GPU memory and \SI{1}{\tera\byte}disk. The external compute payload \cpa\ in Demonstration 2 is an identical Jetson AGX Orin;
\item \textbf{Husarion Panther} host robot \rbb\ in Demonstration 2 with an Intel i9-13900 CPU, \SI{64}{\giga\byte} RAM, \SI{1}{\tera\byte} disk, and no GPU;
\item \textbf{Robotis Turtlebot} host robot \rbc\ in Demonstration 2 with a Raspberry Pi 4 Model B, \SI{2}{\giga\byte} RAM, \SI{32}{\giga\byte} disk, and no GPU.
\end{itemize}
Additional payloads are interfaced with the host robots as described in \Cref{sec:approach}.
Full demonstration videos are available at {\footnotesize\projectsite}.

\begin{figure*}[htbp!]
    \centering
    \vspace*{1ex}
    \includegraphics[width=0.95\textwidth]{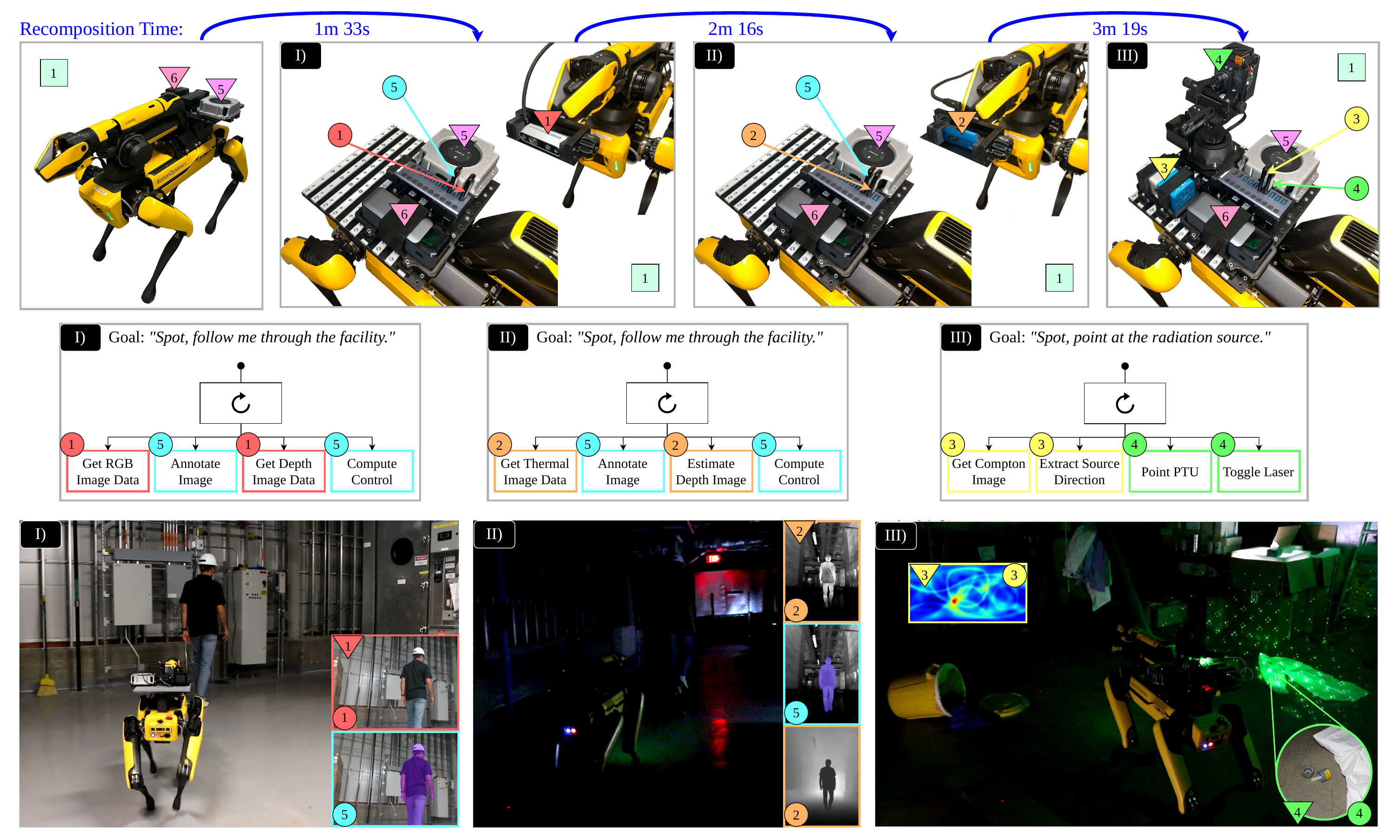}
    % \includegraphics[width=0.95\textwidth]{figures/images/netl_demo.pdf}
    % \caption{The radiation source localization demonstration, consisting of I) person-following ingress into the building using an RGB-D camera, II) person-following traversal inside the dark facility using a thermal camera, and III) radiation source localization and indication with a Compton radiation camera and pan-tilt unit with laser. Top shows the software and hardware payloads plugged into the host robot during each stage. Middle shows the natural language goal provided by the user during each stage and the corresponding autonomously generated task-level behavior trees leveraging the software available at each stage. Bottom shows images from third-person perspective and sensor data captured and processed onboard the robot during execution of each task. Software and hardware labels match Fig. \ref{fig:components} and behaviors and robot data snapshots indicate the components that enable them.}
    \caption{\textbf{Demonstration 1: Radioactive Source Localization} consisting of stage I) person-following traversal through the building using an RGB-D camera, stage II) continued traversal through a dark part of the facility using a thermal camera, and stage III) radiation source localization and indication with a Compton radiation camera and pan-tilt unit with laser. 
    The top row shows the \unseen payloads plugged into the host robot during each stage. 
    The middle row shows the user-provided natural language goals during each stage and the corresponding autonomously generated task-level behavior trees. 
    The bottom row shows images from third-person perspective and onboard sensor data collected and processed during task execution. 
    Software and hardware tags match \Cref{fig:components} and behaviors and robot data snapshots indicate the components that enable them.
    Recomposition time for each stage is shown above the top row.}
    \label{fig:netl_demo}
    \vspace*{-10pt}
\end{figure*}

\subsection{Demonstration 1: Radioactive Source Localization}\label{demo:subsection:netl_demo}

\textbf{Scenario:}
We mimic a partial blackout at an operational nuclear reactor during which an internal radiation alarm has been triggered. 
Demonstrators placed actual radioactive material in a location that was out of plain sight, but from which beta and gamma emissions would be detectable by instruments.
A Spot host robot is deployed alongside a human teammate with a suite of \unseen payloads to be used for person tracking and radioactive source localization.
This demonstration is summarized in \Cref{fig:netl_demo}.

\textbf{Execution:}
Spot \rba\ is initially configured with software components shown in \Cref{fig:components} and a speaker and microphone \hwe\ and external battery \hwf. 
To navigate through the building without a prior map, the robot can follow the teammate. To enable this, the teammate plugs in an RGB-D camera \hwa\ with supporting driver software \swa\ and a YOLO image detection model and vision-based tracking controller \swe.
The teammate issues a command to \textit{``follow me through the facility''}, and the system-level behavior tree, shown in \Cref{fig:system_diagram}, generates a task-level behavior tree for person-following enabled by the added payloads, enabling the robot to follow the teammate through the building.
This stage is shown in \Cref{fig:netl_demo}-I.

\begin{figure*}[htbp!]
    \centering
    \vspace*{1ex}
    \includegraphics[width=0.95\textwidth]{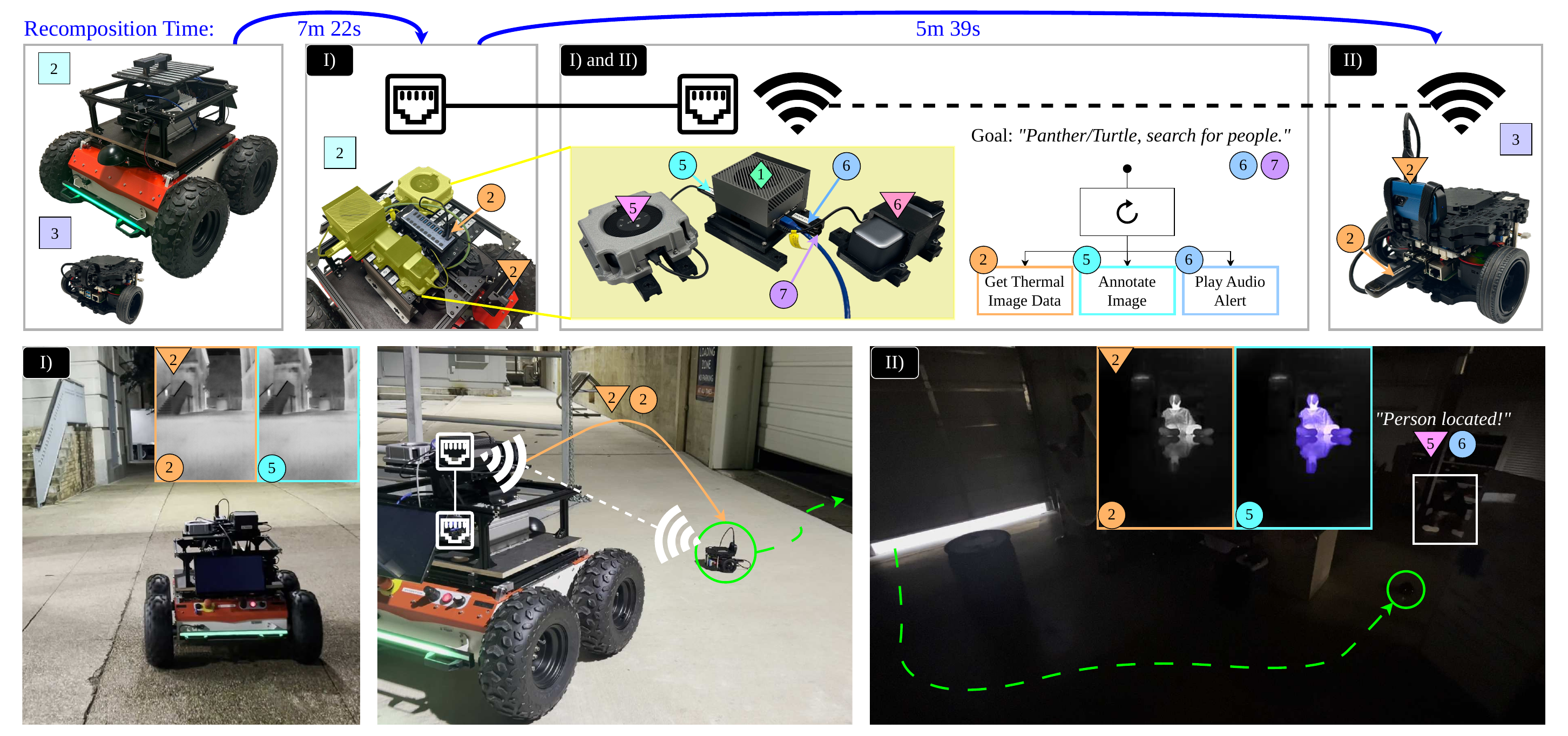}
    % \includegraphics[width=0.95\textwidth]{figures/images/ahg_demo.pdf}
    % \caption{\textbf{Demonstration 2: Person Search in Darkness}: 
    % The person search demonstration, consisting of I) an initial outdoor survey with a rugged robot before II) transitioning to a much smaller robot capable of entering a dark building through a small opening. 
    % Both robots are computationally limited, so an external compute payload is supplied to run required computationally expensive processes. 
    % This compute payload is connected via ethernet with the larger robot and wirelessly with the small robot. Top shows the software and hardware payloads integrated with each host robot and the external compute payload during each stage. Middle shows the natural language goals provided to each robot, the corresponding autonomously generated task-level behavior trees, and snapshots of sensor data collected during operation. Bottom shows third-person images captured during each stage. Software, hardware, and compute labels match Fig. \ref{fig:components} and behavior and robot data snapshots indicate the components that enable them.}
    \caption{\textbf{Demonstration 2: Thermal-Guided Person Search} consisting of stage I) an outdoor survey with a Panther robot and stage II) the transition to a Turtlebot robot capable of entering a dark building through a small opening. 
    Both robots are initially computationally limited and augmented with the same external compute payload to run GPU-intensive processes. 
    The external compute payload is connected via Ethernet with the Panther and wirelessly with the Turtlebot.
    The top row shows payloads integrated with each host robot, including the shared external compute payload alongside the natural language goal provided to both robots and the corresponding autonomously generated task-level behavior tree.
    The bottom row shows third-person images and onboard sensor data captured during each stage. 
    Software, hardware, and compute payload tags match \Cref{fig:components} and behavior and robot data snapshots indicate the components that enable them.
    Recomposition time for each stage is shown above the top row.}
    \label{fig:ahg_demo}
    \vspace*{-10pt}
\end{figure*}

During ingress, a blackout kills all lights and the RGB-D camera becomes ineffective.
To adapt to this change in environment, the teammate replaces the camera and drivers (\hwa, \swa) with a thermal camera \hwb\ plus supporting drivers and a monocular depth-estimation model \cite{depth_pro_2024} \swb.
Once the new payloads are loaded, the teammate provides the same goal \textit{``follow me through the facility''}.
The new thermal sensing modality is composed with the already-running image detection model and tracking controller to continue moving through the building.
This stage is shown in \Cref{fig:netl_demo}-II.

Finally, once arriving at the contaminated room, the teammate removes the running software and hardware payloads (\swb, \hwb, \swe) and plugs in a Compton radiation camera \hwc\ with corresponding drivers \swc\ and a pan–tilt unit with laser \hwd\ and corresponding drivers \swd.
The teammate provides the goal \textit{``point at the radiation source''}, and the system-level behavior tree generates a task-level behavior tree to detect radiation with the Compton camera and point to it with the PTU laser, successfully localizing the hidden source.
This stage is shown in \Cref{fig:netl_demo}-III.

% \textbf{Key Result:}
% This scenario validates \ref{attribute:A} incremental payload integration during execution. Throughout the mission, payloads were swapped (\hwa, \swa, \swe $\to$ \hwb, \swb, \swe $\to$ \hwc, \swc, \hwd, \swd) to adapt to changing environmental conditions and mission objectives. 
% \redtext{To save space we can cut this and point at \Cref{fig:netl_demo}}
% Recomposition for stage I) took 1 minute and 33 seconds including physical mounting and software installation.
% Subsequent recomposition for stage II) took 2 minutes and 16 seconds, and the final recomposition for stage III) took 3 minutes and 19 seconds.
Recomposition times between stages, including the teammate physically interfacing new payloads and automatic discovery and loading of all resources, are provided in \Cref{fig:netl_demo}.

\subsection{Demonstration 2: Thermal-Guided Person Search}\label{demo:subsection:ahg_demo}

\textbf{Scenario:}
We evaluate our framework in a mock disaster-search task, shown in \Cref{fig:ahg_demo}, where a large Panther host robot is initially configured to search for people outdoors using a thermal camera (\Cref{fig:ahg_demo}-I) but is too large to physically fit through a discovered confined building entryway.
A smaller Turtlebot host robot is powered on and, utilizing resources shared from the external compute onboard the Panther,  enters the building and continues the search (\Cref{fig:ahg_demo}-II).

\textbf{Execution:}
The Panther \rbb\ with initial software components shown in \Cref{fig:components} is deployed to search for signs of people outside the building. 
A GPU-equipped external compute payload \cpa\ is connected to the Panther via Ethernet to support GPU-intensive software required for this task.
The \texttt{Component Manager} instances on both the Panther and external compute payload automatically discover each other.
The external compute payload brings its own power \hwf\ and interfaces with the same speaker and microphone \hwe\ from Demonstration 1.
Software payloads containing a local speech-to-text model \cite{whisper_2023} and text-to-speech API \swf\ and a local LLM \cite{llama3_2024} \swg\ are plugged into the external compute payload and all associated databases are forwarded to the Panther.
The same image detection model and tracking controller \swe\ from Demonstration 1 is plugged into the external compute payload and its databases are forwarded to the Panther.
Finally, the same thermal camera \hwb\ and supporting software \swb\ from Demonstration 1 are plugged directly into the Panther. 
Once all software has loaded, the teammate provides the goal \textit{``search for people''} and the system-level behavior tree generates and executes a task-level behavior tree to alert if any people are detected, utilizing components from both the Panther and the external compute payload.
This stage is shown in \Cref{fig:ahg_demo}-I.

During the mission, the Panther encounters a narrow opening into the affected building.
To explore inside the building, a smaller, modified Turtlebot robot \rbc\ is powered on.
The \texttt{Component Manager} instances on the Turtlebot and external compute payload discover each other through a shared wireless network, and all active component databases from the compute payload are forwarded to the Turtlebot.

The thermal camera and drivers (\hwb, \swb) are removed from the Panther and plugged into the Turtlebot.
Once all payloads are loaded, the teammate provides the Turtlebot the same natural language goal and the Turtlebot generates and executes the same task-level behavior tree using the same components on the external compute payload and the transferred thermal camera payloads from the Panther.
The Turtlebot enters the dark interior of the building through the small opening and proceeds until it detects a person, alerting the teammate outside using the text-to-speech software and speaker hardware connected to the external compute payload. 
This stage is shown in Fig. \ref{fig:ahg_demo}-II.

% \textbf{Key Result:}
% This demonstration validates \ref{attribute:B} cross-system capability sharing. 
% To overcome computational limitations of the Panther \rbb\ and Turtlebot \rbc, both robots seamlessly integrated resources made available (\hwe, \swe, \swf, \swg) by the external compute payload \cpa.
% \redtext{To save space we can cut this and point at \Cref{fig:ahg_demo}}
% Recomposition for stage I), including both the teammate physically interfacing new payloads to the Panther and external compute payload and the automatic discovery and loading of all resources between both systems, took 7 minutes and 22 seconds.
% Recomposition for stage II), including discovery, sharing of resources running on the external compute payload, and transferring the thermal camera and supporting software to the Turtlebot, took 5 minutes and 39 seconds. 
Recomposition times between stages, including the teammate physically interfacing new payloads and automatic discovery and loading of all resources, are provided in \Cref{fig:ahg_demo}.

\subsection{Outcomes}
\begin{figure}[t!]
    \vspace{1ex}
    \centering
    
    \begin{minipage}[t]{0.22\textwidth}
        \vspace{0pt}
        \centering
        \includegraphics[width=\textwidth, trim=0.75cm 1cm 0.75cm 0cm, clip]{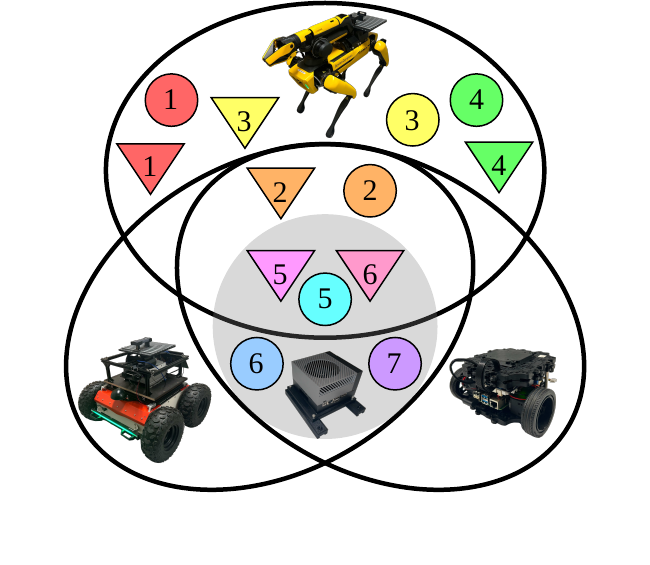}
        {\footnotesize (a)}
        \label{fig:venn}
    \end{minipage}
    % \hfill
    \begin{minipage}[t]{0.25\textwidth}
        \vspace{15pt}
        \centering

        {\small
        \setlength{\tabcolsep}{2pt}    % Narrower columns
        \begin{tabular}{@{}lccc@{}}
        \toprule
        \textbf{Attribute}  
        & \textbf{D1} 
        & \textbf{D2} 
        % & \textbf{D1+D2} 
        & \textbf{\begin{tabular}[c]{@{}c@{}} D1+ \\ D2\end{tabular}}
        \\ \midrule
        
        \textbf{\ref{attribute:A}: Incremental} 
        % \textbf{\ref{attribute:A}} 
        & \checkmark & & \checkmark \\
        
        \textbf{\ref{attribute:B}: Distributed}
        % \textbf{\ref{attribute:B}}
        & & \checkmark & \checkmark \\ 
        
        \textbf{\ref{attribute:C}: Host-Agnostic} 
        % \textbf{\ref{attribute:C}} 
        & \checkmark & \checkmark & \checkmark \\
        
        % \textbf{A4: \begin{tabular}[t]{@{}l@{}} Concert of \\ Ind. Payloads\end{tabular}}
        % \textbf{A4: \begin{tabular}[t]{@{}l@{}} Concert of \\ Payloads\end{tabular}}
        \textbf{\ref{attribute:D}: Concert}
        % \textbf{\ref{attribute:D}}
        & \checkmark & \checkmark & \checkmark \\ 
        \bottomrule
        \end{tabular}
        }
        \par\vspace{20pt}
        {\footnotesize (b)}
        \label{tab:comparison}

    \end{minipage}
    \caption{\textbf{Demonstration Outcomes:} 
    (a) Venn diagram showing the reuse of hardware, software, and compute payloads across host robots. 
    For robots that deployed the external compute payload (\rbb\,, \rbc) the payloads deployed on the external compute are shaded gray.
    % Each hardware payload shown also requires supporting software shown in \Cref{fig:components}, e.g. \swa, \swb, \swc, \swd. 
    % (b) A summary of which attributes introduced in \Cref{sec:demonstrations} are validated in each demonstration (\textbf{D1: Demo 1, D2: Demo 2}). All attributes are collectively validated between the demonstrations.
    (b) Validation of plug-and-play recomposition during runtime attributes introduced in \Cref{sec:demonstrations} across Demonstrations 1 and 2 (\textbf{D1, D2}); all attributes are collectively validated between the demonstrations.}
    \label{fig:outcomes}
    \vspace{-10pt}
\end{figure}
Our demonstrations show rapid payload utilization and reuse across host robots allowed by our core contributions: composable abstractions and a complementary plug-and-play framework. 
Collective outcomes are summarized in \Cref{fig:outcomes}.
\Cref{fig:outcomes}(a) shows the significant overlap between payloads from \Cref{fig:components} used across all robots during both demonstrations, despite the differences in morphology, computational resources, environment, and task between systems and demonstrations.
\Cref{fig:outcomes}(b) summarizes the alignment between demonstrations and attributes introduced in \Cref{sec:demonstrations}. 
Demonstration 1 validates \ref{attribute:A} with incremental modifications to the robot based on changes in task and environment, Demonstration 2 validates \ref{attribute:B} with sharing of resources between robots and external compute payloads, and both demonstrations validate \ref{attribute:C} and \ref{attribute:D} with reusing identical payloads between systems across environments and tasks and coordinating several independent payloads together to perform those tasks.
% demonstrate the framework's efficacy across a combination of host robots and payloads.
% The bolded entries highlight the reuse of payloads between Spot, Panther, and Turtlebot.
% The final attribute, \ref{attribute:D}, is validated by the task-level behavior trees composed from independently design payloads.

% Together, these two demonstrations validate the remaining two core dimensions of our framework: \ref{attribute:C} host-agnostic integration and \ref{attribute:D} the concert of independently designed payloads.
% Across both demonstrations, the same thermal camera \hwb\ and driver software \swb\ were used on the three host robots (\rba, \rbb, and \rbc), while the person detection software \swe\ was used on both Spot \rba\ and the external compute payload \cpa in different tasks and environments.
% Further, in both scenarios, multiple independently developed payloads--ranging from thermal imaging to radiation sensing to natural language interaction--were composed into cohesive, executable behavior trees in response to provided goals. 

% Most importantly, our framework reduced the time required to reconfigure a robot to perform a new task using previously unseen components to just a few minutes. 
% In contrast, traditional methods would likely involve terminating running processes, installing new libraries, resolving dependencies, rewriting code, modifying configuration files, and restarting the system, potentially requiring hours or even days of expert effort. 

As shown in both demonstrations, our framework enables non-roboticist end users to perform reconfiguration tasks that are normally inaccessible without domain expertise.
Traditional reintegration requires network access, credentials, and system-level familiarity before entering the standard robotics development cycle of modifying, tuning, and testing. Even for experts, this process can lead to system downtime orders of magnitude greater than the minutes-scale turnaround enabled by our approach \cite{robot_software_reconfiguration_2025,challenges_of_robot_testing_2020}, which represents a significant improvement toward creating robots that can flexibly adapt within dynamic environments.
% Especially in timely and high pressure applications, our ability to reduce system downtime to minutes and remove the integration burden from the operator represents a significant improvement over existing methods \cite{german_task_force_lessons_2024}.

% With our framework, the human effort required for reconfiguration was limited to physically swapping devices and issuing high-level commands, while all system-level updates were handled automatically.
\section{Limitations}\label{sec:limitations}

% As noted in \Cref{sec:introduction}, we focus on the scheduling, information flow, and utilization of dynamically introduced payloads.
% Other pertinent considerations include the selection of components based on task-alignment or fitness within a particular environment, which ultimately remain the responsibility of the human to handle.

% It ultimately remains up to the operator to provide the robot with the components it needs to perform the desired task, and the robot will subsequently try to generate and execute a task-aligned plan using its available components.

% As noted in \Cref{subsec:approach:component_design_abstractions}, hardware mounts are also not currently proprioceptive, so the operator must provide information about the relative pose of newly introduced hardware to the robot.

% Use of PDDL as an intermediate stage in behavior tree generation is another limitation, as the inability of PDDL to generate plans with control structures more complicated than sequences and parallel execution prevent us from capitalizing on the reactivity possible with behavior trees.
% An alternative behavior tree generation mechanism that can take advantage of these advantages would be highly beneficial.
% \hlgreen{format this independently for both highlighted and regular versions}
As noted in \Cref{sec:introduction}, our scope focuses on the scheduling, information flow, and utilization of dynamically introduced payloads.
While the framework enables runtime reintegration and task-level reasoning over newly available components, several important aspects remain outside its scope. 
% First, the system does not autonomously determine which payloads should be deployed for a given task or environment.
First, payload selection, including task alignment and environment suitability, remains the responsibility of the teammate.
At planning time, the robot assumes it is equipped with a payload suite sufficient to accomplish the provided goal. 
Second, 
% as noted in \Cref{subsec:approach:component_design_abstractions},
hardware mounts are not currently proprioceptive, implying there no is verification of sensor or actuator placement, field of view, or utility.
Teammates must therefore ensure physical payloads are mounted in appropriate configurations.
% and manually provide relative pose information.
% Considerations such as sensor placement, field of view, or downstream effects of physical configuration are not autonomously validated. 
% Third, our current task-planning pipeline uses PDDL as an intermediate representation for behavior tree generation.
% Classical PDDL planners produce non-reactive, primarily sequential and parallel plans.
% As a result, the full expressive and reactive potential of behavior trees is not realized.
% A behavior-tree-native synthesis mechanism would enable generation of more sophisticated task plans.
Third, because we use a classical PDDL planner that produces non-reactive, sequential and parallel plans, the full expressive and reactive potential of behavior trees is not yet realized.
% This is a limitation of planning, not representation, and a
A capable behavior-tree-native synthesis mechanism would enable reactive plan generation using \unseen payloads.
% A behavior-tree-native synthesis mechanism would therefore be required to enable more sophisticated task plans.
\section{Conclusions}

% The state of a robot at initial deployment need not determine its ultimate capabilities. 
We introduce composable abstractions that allow software, hardware, and compute payloads to be dynamically introduced and removed during runtime and a framework that supports recomposition via truly plug-and-play modules and distributed capability sharing.
% We introduce composable abstractions that allow software, hardware, and compute payloads to be dynamically introduced and removed during runtime and a framework that supports both single-system recomposition and distributed systems with seamless capability sharing.
Through demonstrations in radioactive source localization and thermal-guided person search scenarios, we show how the integration of \unseen payloads rapidly expands a robot's operational capabilities to adapt to changing mission requirements on the timescale of minutes.
% Using our approach, system reconfiguration time is reduced to minutes, with users only responsible for physically swapping payloads and issuing high-level goals.
% \hlgreen{
% % Current limitations include proprioceptive sensing of mounted hardware locations on the host robot.
% Two main limitations include: 
% 1. The current system relies on human-guided physical integration; operators must ensure that payloads are mounted in orientations consistent with their intended use.
% 2. Our current framework assumes payload sufficiency for requested tasks, e.g. it is up to the operator to equip the robot with a payload suite that is capable of fulfilling a mission.
% } 
% A future improvement is shifting away from PDDL--which generally creates non-reactive, sequential task plans--toward a generalized task-level behavior tree generation method to produce more complex control structures.
Ultimately, this work challenges the standard robot integration process, demonstrating how with careful abstraction, robots can instead become dynamic, recomposable systems capable of rapidly adapting to emergent mission requirements.
% \input{7_acknowledgment}

%%%%%%%%%%%%%%%%%%%%%%%%%%%%%%%%%%%%%%%%%%%%%%%%%%%%%%%%%%%%%%%%%%%%%%%%%%%%%%%%
% \section*{Acknowledgment}
% We thank Andy Bola\~nos, Zarko Vukovich, Rohan Polavaram, Arjun Nair, and Ayan Chaudhry for their help in performing the demonstrations in this paper. We also thank Bill Charlton, Will Flanagan, and Don Nolting at the UT Nuclear Engineering Teaching Laboratory (NETL) for access to their space and help with the radioactive source localization demonstration.


\printbibliography

@inproceedings{whisper_2023,
    title={Robust speech recognition via large-scale weak supervision},
    author={Radford, Alec and Kim, Jong Wook and Xu, Tao and Brockman, Greg and McLeavey, Christine and Sutskever, Ilya},
    booktitle={Int. Conf. Mach. Learn.},
    year={2023},
}

@article{llama3_2024,
    title={The {Llama} 3 herd of models},
    author={Grattafiori, Aaron and Dubey, Abhimanyu and Jauhri, Abhinav and Pandey, Abhinav and Kadian, Abhishek and Al-Dahle, Ahmad and Letman, Aiesha and Mathur, Akhil and Schelten, Alan and Vaughan, Alex and others},
    journal={arXiv preprint arXiv:2407.21783},
    year={2024}
}

@inproceedings{plansys2_2021,
    title={{PlanSys2}: A planning system framework for {ROS2}},
    author={Mart{\'\i}n, Francisco and Clavero, Jonatan Gin{\'e}s and Matell{\'a}n, Vicente and Rodr{\'\i}guez, Francisco J},
    booktitle={Proc. IEEE/RSJ Int. Conf. Intell. Robots Syst.},
    year={2021},
}

@article{depth_pro_2024,
    title={Depth {Pro}: Sharp monocular metric depth in less than a second},
    author={Bochkovskii, Aleksei and Delaunoy, Ama{\~A}{\c{G}}l and Germain, Hugo and Santos, Marcel and Zhou, Yichao and Richter, Stephan R and Koltun, Vladlen},
    journal={arXiv preprint arXiv:2410.02073},
    year={2024}
}

@article{docker_2014,
    title={{Docker: Lightweight Linux containers for consistent development and deployment}},
    author={Merkel, Dirk},
    journal={Linux J.},
    volume={239},
    number={2},
    year={2014}
}

@article{ros2_2022,
    author = {Steven Macenski  and Tully Foote  and Brian Gerkey  and Chris Lalancette  and William Woodall },
    title = {Robot {Operating} {System} 2: Design, architecture, and uses in the wild},
    journal = {Sci. Robot.},
    volume = {7},
    number = {66},
    year = {2022},
}

@article{bts_collendanchise_2016,
    title={How behavior trees modularize hybrid control systems and generalize sequential behavior compositions, the subsumption architecture, and decision trees},
    author={Colledanchise, Michele and {\"O}gren, Petter},
    journal={IEEE Trans. Robot.},
    volume={33},
    number={2},
    year={2016},
}

@techreport{pddl_1998,
  author={Ghallab, Malik and Howe, Adele and Knoblock, Craig and McDermott, Drew and Ram, Ashwin and Veloso, Manuela and Weld, Daniel and Wilkins, David},
  title       = {{PDDL}-The Planning Domain Definition Language},
  institution = {Yale Center for Computational Vision and Control},
  year        = {1998},
  number      = {CVC TR-98-003/DCS TR-1165},
}

@inproceedings{coral_2026,
    author={Swanbeck, Steven and Pryor, Mitch},
    booktitle={Proc. IEEE/SICE Int. Symp. Syst. Integration}, 
    title={{CORAL}: A Unifying Abstraction Layer for Compositional Robotics Software}, 
    year={2026},
}

@article{citizen_developer_framework_2022,
    title={A framework for rapid robotic application development for citizen developers},
    author={Panayiotou, Konstantinos and Tsardoulias, Emmanouil and Zolotas, Christoforos and Symeonidis, Andreas L and Petrou, Loukas},
    journal={Software},
    volume={1},
    number={1},
    year={2022},
}

@article{component_based_robot_engineering_p1_2009,
    title={Component-based robotic engineering (part {I})},
    author={Brugali, Davide and Scandurra, Patrizia},
    journal={IEEE Robot. Automat. Mag.},
    volume={16},
    number={4},
    year={2009},
}

@article{robot_software_reconfiguration_2025,
    title={Software reconfiguration in robotics},
    author={Peldszus, Sven and Brugali, Davide and Str{\"u}ber, Daniel and Pelliccione, Patrizio and Berger, Thorsten},
    journal={Empirical Softw. Eng.},
    volume={30},
    number={3},
    year={2025},
}

@article{general_purpose_fms_2023,
    title={Toward general-purpose robots via foundation models: A survey and meta-analysis},
    author={Hu, Yafei and Xie, Quanting and Jain, Vidhi and Francis, Jonathan and Patrikar, Jay and Keetha, Nikhil and Kim, Seungchan and Xie, Yaqi and Zhang, Tianyi and Fang, Hao-Shu and others},
    journal={arXiv preprint arXiv:2312.08782},
    year={2023}
}

@article{adaptive_robotics_2022,
    title={A methodical interpretation of adaptive robotics: Study and reformulation},
    author={Enayati, Amir M Soufi and Zhang, Zengjie and Najjaran, Homayoun},
    journal={Neurocomputing},
    volume={512},
    year={2022},
}

@article{trends_in_reconfigurable_modular_robots_2017,
    title={Current trends in reconfigurable modular robots design},
    author={Brunete, Alberto and Ranganath, Avinash and Segovia, Sergio and De Frutos, Javier Perez and Hernando, Miguel and Gambao, Ernesto},
    journal={Int. J. Adv. Robot. Syst.},
    volume={14},
    number={3},
    year={2017},
    publisher={SAGE Publications Sage UK: London, England}
}

@article{heuss_extendable_2022,
    title={An extendable framework for intelligent and easily configurable skills-based industrial robot applications},
    author={Heuss, Lisa and Gonnermann, Clemens and Reinhart, Gunther},
    journal={Int. J. Adv. Manuf. Technol.},
    volume={120},
    number={9},
    year={2022},
}

@inproceedings{heuss_modular_2019,
  author    = {Heuss, Lisa and Blank, Andreas and Dengler, Sebastian and Zikeli, Georg Lukas and Reinhart, Gunther and Franke, J{\"o}rg},
  title     = {Modular Robot Software Framework for the Intelligent and Flexible Composition of Its Skills},
  booktitle = {IFIP Int. Conf. Adv. Prod. Manage. Syst.},
  % volume    = {566},
  year      = {2019},
  %publisher = {Springer, Cham},
}

@article{reconfigurable_software_2002,
    title={Design of dynamically reconfigurable real-time software using port-based objects},
    author={Stewart, David B. and Volpe, Richard A. and Khosla, Pradeep K.},
    journal={IEEE Trans. Softw. Eng.},
    volume={23},
    number={12},
    year={1997},
}

@article{xbot2_middleware_2023,
  title={The {XBot2} real-time middleware for robotics},
  author={Laurenzi, Arturo and Antonucci, Davide and Tsagarakis, Nikos G and Muratore, Luca},
  journal={Robot. Auton. Syst.},
  volume={163},
  year={2023},
}

@article{murt_middleware_2023,
  title={$\mu${RT}: A lightweight real-time middleware with integrated validation of timing constraints},
  author={Sch{\"o}pping, Thomas and Kenneweg, Svenja and Hesse, Marc and R{\"u}ckert, Ulrich},
  journal={Frontiers Robot. AI},
  volume={10},
  year={2023}
}

@article{cloud_robotics_resources_2021,
    title={Resource allocation and service provisioning in multi-agent cloud robotics: A comprehensive survey},
    author={Afrin, Mahbuba and Jin, Jiong and Rahman, Akhlaqur and Rahman, Ashfaqur and Wan, Jiafu and Hossain, Ekram},
    journal={IEEE Commun. Surv. \& Tut.},
    volume={23},
    number={2},
    year={2021},
}

@INPROCEEDINGS{pdra_2020,
    author={Rossi, Federico and Vaquero, Tiago Stegun and Sanchez-Net, Marc and da Silva, Maíra Saboia and Vander Hook, Joshua},
    booktitle={Proc. IEEE/RSJ Int. Conf. Intell. Robots Syst.}, 
    title={The Pluggable Distributed Resource Allocator ({PDRA}): a Middleware for Distributed Computing in Mobile Robotic Networks}, 
    year={2020},
}

@article{temoto_2022,
    title={{TeMoto}: a software framework for adaptive and dependable robotic autonomy with dynamic resource management},
    author={Valner, Robert and Vunder, Veiko and Aabloo, Alvo and Pryor, Mitch and Kruusam{\"a}e, Karl},
    journal={IEEE Access},
    volume={10},
    year={2022},
}

@book{compositional_thinking_2022,
    author={Censi, Andrea and Lorand, Jonathan and Zardini, Gioele},
    title={Applied Compositional Thinking for Engineers},
    year={2022},
    note={Work in Progress Book}
}

@article{dynamic_software_updates_2023,
    title={Scheduling Dynamic Software Updates in Mobile Robots},
    author={Yaacoub, Ahmed El and Mottola, Luca and Voigt, Thiemo and R{\"u}mmer, Philipp},
    journal={ACM Trans. Embedded Comput. Syst.},
    volume={22},
    number={6},
    year={2023},
}

@article{software_variability_2023,
    title={Software variability in service robotics},
    author={Garc{\'\i}a, Sergio and Str{\"u}ber, Daniel and Brugali, Davide and Di Fava, Alessandro and Pelliccione, Patrizio and Berger, Thorsten},
    journal={Emp. Softw. Eng.},
    volume={28},
    number={2},
    year={2023},
}

@inproceedings{challenges_of_robot_testing_2020,
    title={A study on challenges of testing robotic systems},
    author={Afzal, Afsoon and Le Goues, Claire and Hilton, Michael and Timperley, Christopher Steven},
    booktitle={Proc. IEEE Int. Conf. Softw. Testing, Validation and Verification},
    year={2020},
}

@inproceedings{fogros2_2023,
    title={{FogROS2}: An adaptive platform for cloud and fog robotics using {ROS} 2},
    author={Ichnowski, Jeffrey and Chen, Kaiyuan and Dharmarajan, Karthik and Adebola, Simeon and Danielczuk, Michael and Mayoral-Vilches, V{\'\i}ctor and Jha, Nikhil and Zhan, Hugo and LLontop, Edith and Xu, Derek and others},
    booktitle={Proc. IEEE Int. Conf. Robot. Automat.},
    year={2023},
}

@article{plug_and_produce_collaborative_Schou_2017, 
    title={A plug and produce framework for industrial collaborative robots}, 
    volume={14}, 
    number={4}, 
    journal={Int. J. Adv. Robot. Syst.}, 
    author={Schou, Casper and Madsen, Ole}, 
    year={2017}, 
}

@article{Generic_plug_and_produce_Profanter_2021, 
    title={A Generic Plug \& Produce System Composed of Semantic {OPC UA} Skills}, 
    volume={2}, 
    journal={IEEE Open J. Ind. Electron. Soc.}, 
    author={Profanter, Stefan and Perzylo, Alexander and Rickert, Markus and Knoll, Alois}, 
    year={2021}, 
}

@article{Modular_reconfigurable_mobile_robotics_2012, 
    title={Modular and reconfigurable mobile robotics}, 
    volume={60}, 
    number={12}, 
    journal={Robot. Auton. Syst.}, 
    author={Moubarak, Paul and Ben-Tzvi, Pinhas}, 
    year={2012}, 
}

@article{Chennareddy_Agrawal_Karuppiah_2017, 
    title={Modular Self-Reconfigurable Robotic Systems: A Survey on Hardware Architectures}, 
    journal={J. Robot.}, 
    author={Chennareddy, S. Sankhar Reddy and Agrawal, Anita and Karuppiah, Anupama}, 
    year={2017}, 
}

@article{Liang_Wu_Tu_Lam_2024,
  title={Decoding modular reconfigurable robots: A survey on mechanisms and design},
  author={Liang, Guanqi and Wu, Di and Tu, Yuxiao and Lam, Tin Lun},
  journal={Int. J. Robot. Res.},
  volume={44},
  number={5},
  year={2024},
}

@article{Arai2000PlugProduce,
    title   = {Agile assembly system by ``Plug and Produce''},
    author  = {Arai, Tamio and Aiyama, Yasumichi and Maeda, Yusuke and Sugi, Masao and Ota, Jun},
    journal = {CIRP Annals - Manuf. Technol.},
    year    = {2000},
    volume  = {49},
    number  = {1},
}

@article{thorvald_Grimstad_From_2017, 
    title={The {Thorvald II} Agricultural Robotic System}, 
    volume={6}, 
    number={4}, 
    journal={Robotics}, 
    author={Grimstad, Lars and From, Pål}, 
    year={2017}, 
}

@article{Guri2024HeftyAM,
    title={Hefty: A Modular Reconfigurable Robot for Advancing Robot Manipulation in Agriculture},
    author={Dominic Guri and Moonyoung Lee and Oliver Kroemer and George Kantor},
    journal={arXiv preprint: arXiv:2402.18710},
    year={2024},
}

@article{mars_Xu_Li_2022, 
    title={A modular agricultural robotic system {(MARS)} for precision farming: Concept and implementation}, 
    volume={39}, 
    number={4}, 
    journal={J. Field Robot.}, 
    author={Xu, Rui and Li, Changying}, 
    year={2022}, 
}

@article{modkom_Wiedemann_2024, 
    title={Enhancing Development of Modular Application-Specific Configurable Space Robots}, 
    volume={2716}, 
    number={1}, 
    journal={J. Phys.: Conf. Ser.}, 
    author={Wiedemann, H. and Schilling, M. and Chowdhury, P. and Brinkmann, W. and Kien, I. and Li, J. and Langosz, M. and Michelson, E.}, 
    year={2024}, 
}

@article{concert_rossini_2026,
    title={Concert: a modular reconfigurable robot for construction},
    author={Rossini, Luca and Romiti, Edoardo and Laurenzi, Arturo and Ruscelli, Francesco and Ruzzon, Marco and Covizzi, Luca and Baccelliere, Lorenzo and Carrozzo, Stefano and Terzer, Michael and Magri, Marco and others},
    journal={J. Field Robot.},
    volume={43},
    number={3},
    pages={1332--1362},
    year={2026},
    publisher={Wiley Online Library}
}

\end{document}